# Reconstructing the Forest of Lineage Trees of Diverse Bacterial Communities Using Bio-inspired Image Analysis

Athanasios D. Balomenos, Elias S. Manolakos
Department of Informatics and Telecommunications
National and Kapodistrian University of Athens, Ilissia, Greece
{abalomenos}eliasm@di.uoa.gr

*Abstract*—Cell segmentation and tracking allow us to extract a plethora of cell attributes from bacterial time-lapse cell movies, thus promoting computational modeling and simulation of biological processes down to the single-cell level. However, to analyze successfully complex cell movies, imaging multiple interacting bacterial clones as they grow and merge to generate overcrowded bacterial communities with thousands of cells in the field of view, segmentation results should be near perfect to warrant good tracking results. We introduce here a fully automated closed-loop bio-inspired computational strategy that exploits prior knowledge about the expected structure of a colony's lineage tree to locate and correct segmentation errors in analyzed movie frames. We show that this correction strategy is effective, resulting in improved cell tracking and consequently trustworthy deep colony lineage trees. Our image analysis approach has the unique capability to keep tracking cells even after clonal subpopulations merge in the movie. This enables the reconstruction of the complete *Forest of Lineage Trees* (FLT) representation of evolving multi-clonal bacterial communities. Moreover, the percentage of valid cell trajectories extracted from the image analysis almost doubles after segmentation correction. This plethora of trustworthy data extracted from a complex cell movie analysis enables single-cell analytics as a tool for addressing compelling questions for human health, such as understanding the role of single-cell stochasticity in antibiotics resistance without losing site of the inter-cellular interactions and microenvironment effects that may shape it.

## I. INTRODUCTION

Analysis of time-lapse microscopy videos (also known as "cell movies") is an important tool allowing us to "zoom in" and observe dynamic biological processes at the single-cell resolution level [1]. Recent studies have noted its importance for investigating how stochasticity (biological "noise") affects gene regulation, aspects of cell growth, cell proliferation, etc. [2]. Mathematical models are important to form and test hypotheses for such dynamic phenomena [3]. Time-lapse movies if properly analyzed can provide an abundance of time course data, extremely valuable for mathematical models' calibration and validation. Manual cell counting and tracking across image frames are extremely laborious and error prone. The accurate, fully automated segmentation and tracking of individual cells, as they grow and divide in expanding bacterial colonies, remain major challenges [4]. The problem is becoming a lot more difficult in complex movies with many growing bacterial clones in the field of view that may merge to generate overcrowded (dense) and diverse microbial communities. Therefore, automation strategies are essential before we can add time-lapse image analysis in the arsenal of high throughput methods routinely used for systems biology.

Several software packages exist that support the segmentation and tracking of cells in time-lapse cell movies. Among them, we mention Schnitzcells [5], Oufti [6] and SuperSegger [7]. Schnitzcells [5] segments cells and tracks them from frame to frame using an energy function optimization method [8]. Oufti [6] combines several algorithms developed for medical image analysis and computer vision, including clustering, template-matching, active contours, region growing and level set methods so as to segment and track cells. The more recently introduced SuperSegger [7] combines a modified version of the watershed algorithm with neural networks classification to identify cell boundaries. Moreover, it uses an optimization algorithm with a special cost function to track ("link") corresponding cells in consecutive frames. Schnitzcells and Oufti produce low-quality segmentation results in time-lapse movies with overcrowded microbial communities [9]. Therefore their segmentation results provide no solid input for their tracking algorithms in complex movies. On the other hand, while SuperSegger produces acceptable segmentation results, it fails to track cells when colonies start merging.

We have recently introduced *Bacterial Single Cell Analytics* (BaSCA) [9], a methodology and automated pipeline of image pre-processing, image analysis and machine learning algorithms, to extract knowledge at the colony and single-cell level from time-lapse complex cell movies. BaSCA cell segmentation has been shown to be more accurate than other methods, to remain robust when the image quality of cell movie frames fluctuates, or when cell overcrowding becomes severe. Also, in [10] we introduced an effective cell tracking algorithm inspired by Block Matching Motion Estimation for video compression [11]. Under perfect segmentation (ground truth segmentation dataset provided by experts) our algorithm has been shown to outperform that of Schnitzcells'.

However, in complex cell movies with multiple subpopulations growing and merging to become overcrowded cell communities, even a small number of segmentation errors is enough to upset considerably cell tracking performance leading to colony lineage tree errors as cells divide. It is thus apparent that a near perfect cell segmentation is a prerequisite for obtaining correct and deep lineage trees capturing the dynamics and epigenetics of many cell generations when analyzing complex cell movies. Moreover, the ability to continue building these cell genealogy trees even when cells emanating from different clones are no longer separated in space but merge to form larger bacterial communities is essential, since in many cases we want to be able to "see" what happens as cell subpopulations get into proximity. This important capability will also allow us to extract data and knowledge from inter-species imaging experiments For example, when different bacterial sub-populations grow in the same microenvironment, and it is of interest to keep track of their genealogy while also studying the dynamics of their interactions as their cell members get into proximity and then mingle in the same area. This kind of inter-species cooperation or competition experiments are often used to decipher how antibiotic tolerance emerges [12], an extremely important question for human health [13].

In this work, we build on prior successes and introduce a closed loop bio-inspired computational strategy. We exploit knowledge about the expected structure of a colony's lineage tree to automatically identify problematic areas in the segmented image frames, correct cell segmentation errors, and as a consequence improve cell tracking results and improve the corresponding lineage trees of subpopulations interacting in a complex cell movie. In this way, we reconstruct a trustworthy *forest of lineage trees* for all clones that may grow in parallel in the movie. Improving our cell segmentation and tracking algorithms presented in [9][10] increases the amount of useful information extracted from a complex movie thus forming the basis for the development of an automated and high throughput single-cell micro-environment analytics platform. Besides improving analysis robustness, the proposed bio-inspired segmentation correction method opens up several new capabilities. Specifically, it can track multiple sub-populations in the field of view, while excluding from the analysis selected subpopulations; it enables visualization of single-cell properties over forest subtrees as they evolve in time. In general, it extracts individual-based "big data" which are a prerequisite for developing Individual-based Modeling (IbM) approaches, a new important tool in advancing microbial sciences [14].

In Section II we present the closed-loop image analysis strategy for segmentation correction and indicative results demonstrating the obtained improvements. In Section III we summarize our findings and point to future research directions.

## II. BIO-INSPIRED SEGMENTATION CORRECTION

### A. Preliminaries-Motivation

All the results presented in this paper are based on the analysis of a difficult to handle, yet typical, time-lapse microscopy dataset. This phase contrast "cell movie" (video) is of average quality and depicts the simultaneous growth of 19 bacterial clones (micro-colonies) each one emanating from a single *Salmonella Typhimurium* cell in the first frame. As time progresses (every frame corresponds to 5 min. of the real experiment, and there are 78 frames in the movie) the single-cells proliferate and divide, and the different colonies grow and then start merging to give rise to overcrowded bacterial communities with thousands of cells in the field of view.

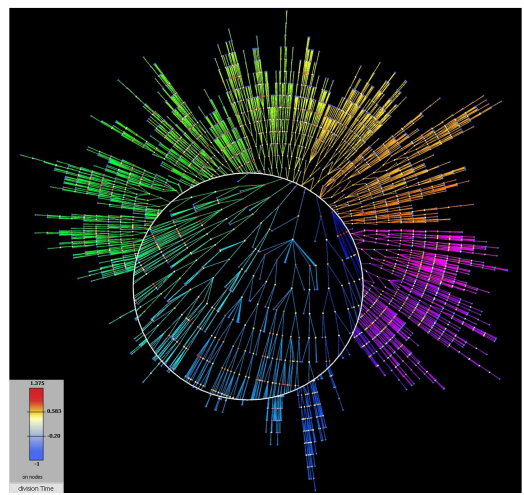

**Fig. 1.** Forest of Division Trees (one colored tree per colony) before segmentation correction. Tree nodes (small circles) mark cell divisions; their color depicts the cell division time (in min.). We observe tree irregularities (see text for details).

The cell movie was analyzed using BaSCA [9]. BaSCA consists of 5 computational stages: image pre-processing, colonies segmentation, single-cells segmentation, cell tracking and lineage trees construction, single-cell attributes estimation and visualization over extracted trees. BaSCA combines image processing with machine learning to detect, segment and characterize each colony and individual cell in every frame of the movie. First, it processes each image frame and extracts individual colony masks. Then it decomposes each colony into a partition of "objects" containing one or more cells. Finally, it zooms in and reaches the desired result; accurate single-cell boundaries detection and cell feature estimation. BaSCA's unique "divide-and-conquer" approach allows the independent analysis of different micro-colonies in the input movie. Its recursive problem decomposition strategy allows to analyze colonies efficiently regardless of their cell density and deal effectively with overcrowded cell images.

Also, in [10] we have presented an effective cell tracking algorithm, inspired by Block Matching Motion Estimation for video compression [11], which can be used to match corresponding cells across consecutive frames, identify cell division events and reconstruct a *Lineage Tree* (LT) structure representation for every growing colony. An LT node represents a cell at a particular time instant (frame) and its color can be used to capture a cell's time varying *attribute* (e.g. cell area). A continuous segment (sequence) of LT nodes between two successive cell divisions represents the lifespan of a cell. New segments are added to the LT when a "mother cell" divides, giving rise to two "daughter cell" segments. If we condense cell segments down to a single node, we obtain the colony's *Division Tree* (DT) capturing only the cell division events. We can use color for each DT node to represent visually single-cell *life attributes* (e.g. cell division times) as a colony grows. Visual analytics on LTs and DTs can be very useful to "see" epigenetic effects from generation to generation, especially if we can maintain the ability to keep building the LT and DT trees even after colonies merge to form large cell communities (e.g. biofilms). Knowing to what original clone each cell belongs at without the need to label cells with molecular markers is very useful for studying interspecies competition, epigenetic correlations, etc. and a unique capability of our image analysis approach that is not supported by any other method.

For example, in Figure 1 we can see the Forest of all Division Trees (FDT) summarizing the analysis of the complex cell movie using BaSCA (segmentation and tracking). We use a different color for each colony's DT. The color of DT nodes (small circles) represents the division time of the cells (in hours). The visualization of the forest was produced using the tool Tulip [15]. Despite the fact that in many areas of the movie the tracking was very good as evidenced by the FDT, we can identify places (see magnified area) where tree structure errors exist, e.g. mother cells having more than two daughters, false cell divisions, cell lineages that stop abruptly, etc. We mainly attribute these problems to segmentation errors that can severely hamper tracking performance. In the sequel, we will show how knowing where these errors occur can help us correct the segmentation locally and thus eliminate them.

*B. Using the expected LT structures to correct Segmentation*

In general, we can use the LT to identify three types of segmentation errors: (i) multiple over-segmentation of a cell in the current frame *t*, (ii) over-segmentation of a cell in one or more previous and consecutive frames, (i.e. frame *t-1, t-2, …*), and (iii) under-segmentation of a cell in the current frame *t*.

In Fig. 2 (b) top, we see how the lineage tree is structured after the tracking when a cell is erroneously over-segmented into multiple fragments. The algorithm tracks this error after matching the cells of current frame *t* with the cells of the previous frame *t-1*. It corrects this error by coalescing multiple cells (red nodes) to generate either one cell (green merged object) or potentially two cells (division event, two sister cells are shown as green merged objects, see Fig. 2 (c) top).

In Fig. 2 (b) bottom, we see an unbalanced tree structure which implies that a segmentation error has occurred since a "lonely" daughter cell (red box) at time *t* is missing its sister cell (indicated by a black dashed box). The algorithm has to determine first if that is due to an under- or over-segmentation error to apply proper correction. If the "lonely" cell has lived at least for three frames (a tunable parameter) the error is first treated as a possible under-segmentation at the current frame *t*. So the algorithm attempts to match the "lonely" red cell with two cells, the one above it in the LT (to which it has been matched) and the other next to it (its sister) that remained unmatched in the current frame triggering the corrective action (see in Fig. 2 the two cells encircled by the red dashed oval). To check the quality of this matching attempt, we apply a modified version of the tracking algorithm presented in [10]. Specifically, we now consider as "previous frame" the current frame *t* and as "current frame" the previous frame *t-1*. So, the tracking algorithm defines a neighborhood based on the two encircled cells in frame *t-1* (to be called $N_{t-1}$) and searches for the best corresponding neighborhood $N_t$ containing the "lonely" cell in frame *t*. We consider that a successful neighborhood correspondence is found if the *total overlap score* (see equation (4) in [11]) of the matchings of cells in frame *t* and cells in neighborhood $N_{t-1}$ exceeds a threshold *T* (currently 75%). If a successful neighborhood correspondence is found the algorithm splits the under-segmented "lonely" cell by using the centroids of the two encircled cells (matched to its area) to initialize the complex object segmentation procedure of BaSCA that performs this task (see Fig. 2 in [9]). Then it updates the tree by also splitting the "lonely" node at level *t* and creating two cells, as shown in Fig. 2 (c) bottom-left. Let us mention that in addition to a sister node, a cousin (or an aunt) of a "lonely" node at level *t* can also go missing… If the missing cousin happens to be included in neighborhood $N_{t-1}$, the tracking algorithm could in one pass assign more than two cells to the under-segmented "lonely" cell. A case like this is shown in Fig. 2 (a) right column where three cells were erroneously considered as one (middle-right). The auto correction scheme identified the inconsistency at the tree level and split the under-segmented cell recovering the three discrete cells. Then it updated the two sibling branches and the extra branch (of the found cousin or aunt).

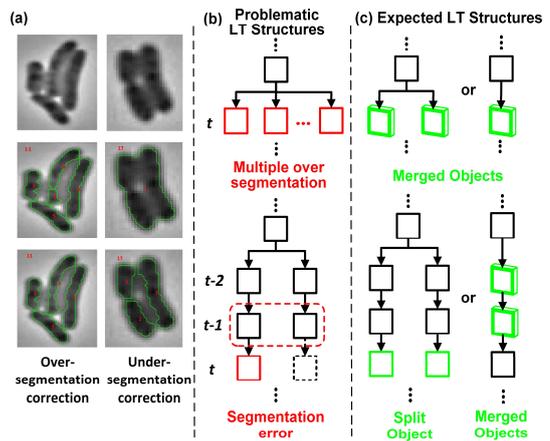

**Fig. 2.** Automatic Correction Scheme. **(a)** *From top to bottom*: input image, erroneous segmentation, corrected segmentation. *Left:* Over-segmentation, *Right:* under-segmentation. **(b)** Problematic LT structures triggering corrective actions. *Top:* Multiple over-segmentation case (1 mother cell matched to *n>2* daughter cells (red nodes); multiple objects are merged resulting in either one cell (panel c top-left) or two cells (panel c top-right green nodes). *Bottom:* LT structure when a segmentation error occurs at frame *t* creating a "lonely" cell (red) with no sister (missing sister indicated by dashed black box). If it is due to a previous over-segmentation error, the sister nodes are merged recursively up until the mother cell is reached (panel c bottom-left). If it is an under-segmentation error at the current frame, the "lonely" node is split into multiple nodes (panel c bottom-right) so as to keep the tree consistent. In both cases, the cell boundaries are modified accordingly to match the tree structure corrections (see text for details). **(c)** Corrected lineage tree (LT) structures.

If the neighborhood matching is not successful, the algorithm considers the error as being an over-segmentation that occurred at previous frames and attempts to correct it by merging pairs of nodes (cells) recursively at the preceding LT levels up and until the last division event (mother cell) is reached. This objects merging process may be aborted at any level, either because the object boundaries of the sister nodes to be merged do not touch i.e. they clearly correspond to distinct cells, or because the overlap score of the newly formed cell (say at frame *t-1*) and the cell at frame *t* is under a set threshold *M* (currently 90%). If merging is successful, the algorithm corrects the cell boundaries and the tree structure (merging sister nodes) so as to keep the tree consistent (see Fig. 2 (c) bottom-right).

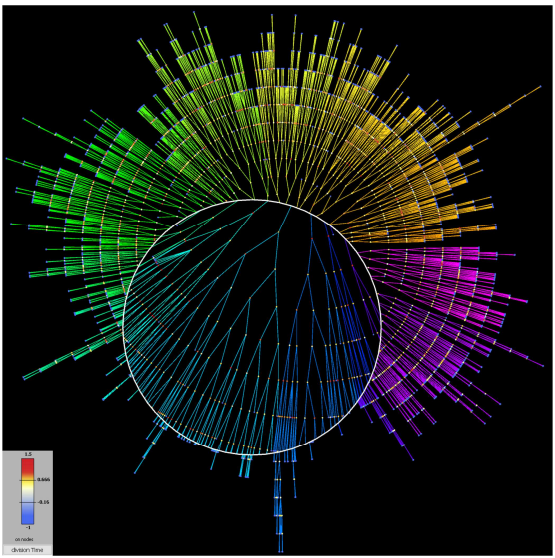

**Fig. 3.** Forest of Division Trees (one colored DT per colony) after applying segmentation correction using the proposed method. DT nodes (small circles) mark cell divisions; their color depicts the cell division times (in min.). We observe a mostly regular binary tree structure (see magnified area) as opposed to the FDT of Figure 1.

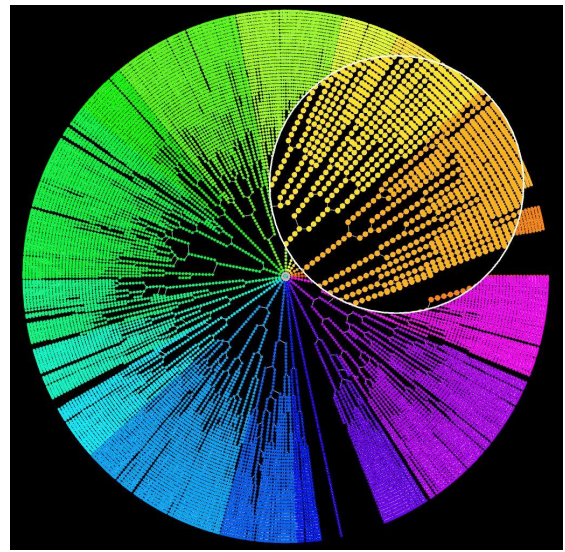

**Fig. 4.** Forest of Lineage Trees after segmentation correction. Colors are used for the LTs of different clonal subpopulations (19 in total) that emanated from single-cells initially and are growing simultaneously in the cell movie. The LTs are 78 levels deep (total number of frames in the processed cell movie). We can observe in the magnified area that segmentation auto correction restored balance leading to valid cell segments (see text for details).

In extreme cases, it is possible that both aforementioned correction actions may fail. In this case, the algorithm keeps the irregular LT structure and continues to the next possible correction. It is worth mentioning that even though our approach is based on the expectation that a correct LT will be balanced, it can still handle a potential cell death event leading to a tree branch that stops growing prematurely.

### C. The Effects of Improved Segmentation on Tracking

In Figure 3 we present the Forest of Division Trees (FDT) for the same movie but after applying segmentation correction based on the approach presented above. When compared to the FDT of Figure 1 (before correction) we observe the following important improvements: the individual colony DTs are more consistent, mother nodes when they divide to give rise to only two daughter cells as expected, the forest is more balanced, as expected in a dataset where cell death is rare.

The Forest of Lineage Trees (FLT) for the same movie after applying segmentation correction is provided in Figures 4. In the magnified section we observe that the LTs are balanced and have valid branches. The correction of segmentation errors led to improved tracking and consequently to more balanced lineage tree structures for the yellow and light orange colonies.

The percentage of valid cell segments in the FLT increased from 46% to 89% of the total number of extracted cell segments, by applying segmentation correction. Moreover the total number of valid cell segments increased by 25% relatively to the non-corrected FLT. A cell segment (cell trajectory) is a sequence of time points in the LT and is considered valid if it adequately represents the lifespan of a cell, i.e. if it starts and ends at two successive cell division points. Cell segments that may end (start) at a non-cell division time point are considered valid only if this point corresponds to the last (first) frame of the movie, i.e. it is actually at a leaf (root) respectively of a colony's LT in the forest (see magnified area of Figure 4).

In Figure 5 (a) we see a part of frame 56 of the movie after segmentation correction and tracking. Each extracted single-cell is colored according to the clone from which it emanated. The corresponding corrected Forest of Lineage Trees (up to frame 56) for the depicted colonies is provided in Fig. 5 (c). We observe here that some initial subpopulations have already merged at that point to form larger communities, and the algorithm keeps the trees of each population consistent. As subpopulation merging may happen at any time in a complex movie, the need for having an algorithm that can continue tracking cells consistently when that happens is apparent. The same information but for the later frame 77 in the movie is provided in Figures 5 (b) and (d) respectively. Despite the fact that many colonies have merged in the time period between the two frames, tracking continues successfully, and as the LTs are expanding, they are used to improve cell segmentation and then improve cell tracking even more. The gray cells belong to colonies that are partially in the field of view (not tracked) or to cells that the algorithm failed to track.

### III. CONCLUSIONS

We have introduced a bio-inspired auto correction strategy that can fix segmentation errors and consequently improve cell tracking and increase significantly the percentage of valid cell segments that can be extracted by the image analysis of complex time-lapse bacterial cell movies. Our computation strategy enables the effective spatial and temporal analysis of overcrowded communities, with multiple and merging subpopulations, both at the colonies and single-cell levels.

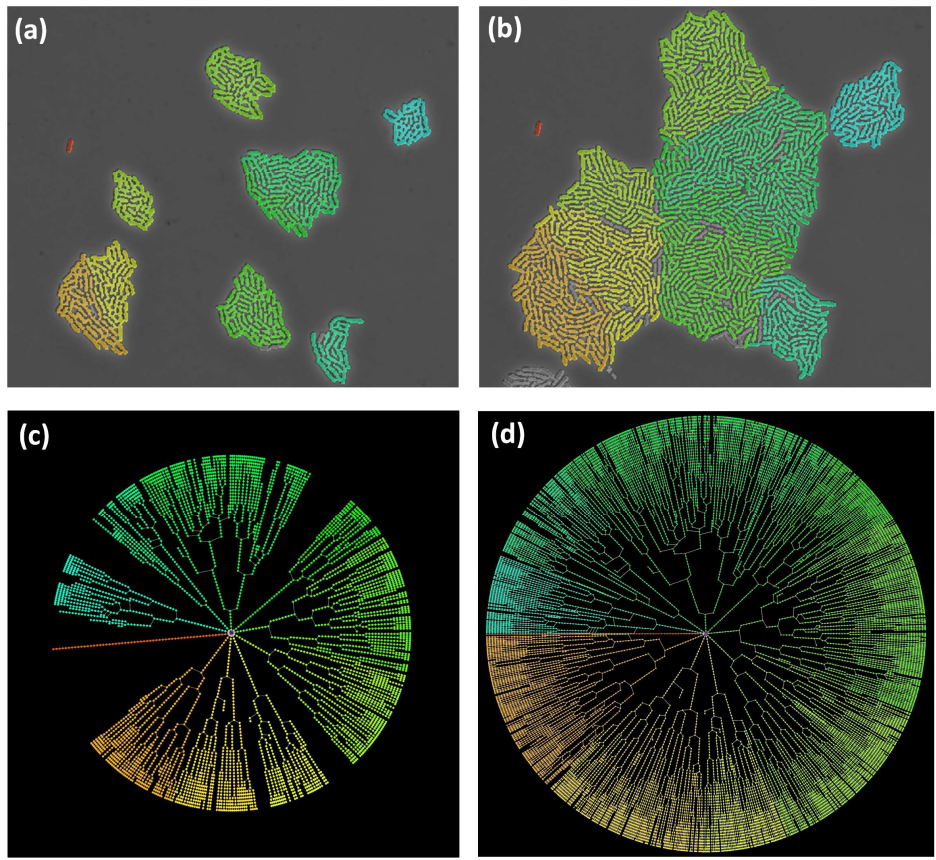

**Fig. 5.** Tracking cell subpopulations as they merge without using cell labeling; the different colors are used here just to visualize the 10 clonal subpopulations growing in the bacterial cell movie. **(a)** State of clonal subpopulations (at frame 56). **(b)** The final state of clonal merged subpopulations (at frame 77). **(c)** and **(d)** The corresponding Forest of Lineage Trees; initially, the trees were sparse and then they become denser as the subpopulations grow and merge. The proposed algorithm continues to track subpopulations after they merge and keeps building the forest of cell subpopulations.

We can track and record accurately how colony and cell morphological and expression characteristics evolve in space and with time and track division events without the need to tag molecularly the different bacterial subpopulations that coexist. To the best of our knowledge, there is currently no other method that can achieve label-free tracking of bacterial subpopulations even after they merge. Moreover, the distribution of extracted single-cell attributes can be mapped and visualized instantly over the forests of lineage or division trees to help scientists see the whole picture (community state) as they try to formulate new hypotheses for further experimental or modeling work.

ACKNOWLEDGMENTS

The first author acknowledges the support of the Alexander S. Onassis Public Benefit Foundation (Scholarship number GZJ030/2013-2014).